\documentclass{svproc}
\usepackage{url}

\usepackage{graphicx}
\usepackage{amsmath,amssymb} 
\usepackage{color}
\usepackage{times}
\usepackage{epsfig}
\usepackage[utf8x]{inputenc}
\usepackage{mathtools}
\usepackage{makecell}
\usepackage{numprint}
\npthousandsep{\,}
\usepackage{algorithm}
\usepackage{multirow}
\usepackage{pifont}
\usepackage[width=122mm,left=12mm,paperwidth=146mm,height=193mm,top=12mm,paperheight=217mm]{geometry}
\newcommand{\xmark}{\ding{55}}%

\newcommand*{\suchthat}[1]{\left|\vphantom{#1}\right.}
\DeclareMathOperator*{\argmin}{arg\,min} 

\newcommand{\etal}{\textit{et al}.}

\newcommand{\eg}{\textit{e}.\textit{g}.}

\usepackage[breaklinks=true,bookmarks=false]{hyperref}

\begin{document}
\mainmatter              
\title{A Dense-Depth Representation for VLAD descriptors in Content-Based Image Retrieval}

%
%
\author{Federico Magliani \and Tomaso Fontanini \and Andrea Prati}

\authorrunning{ISVC-18 submission ID 7}


%
\institute{IMP lab, D.I.A. - University of Parma, 43124,  Parma, Italy\\
\email{federico.magliani@studenti.unipr.it},\\
\texttt{http://implab.ce.unipr.it}
}
\maketitle              

\begin{abstract}
The recent advances brought by deep learning allowed to improve the performance in image retrieval tasks. Through the many convolutional layers, available in a Convolutional Neural Network (CNN), it is possible to obtain a hierarchy of features from the evaluated image. At every step, the patches extracted are smaller than the previous levels and more representative. Following this idea, this paper introduces a new detector applied on the feature maps extracted from pre-trained CNN. Specifically, this approach lets to increase the number of features in order to increase the performance of the aggregation algorithms like the most famous and used VLAD embedding.
The proposed approach is tested on different public datasets: Holidays, Oxford5k, Paris6k and UKB.
\end{abstract}

\section{Introduction}

The recent growth of available images and videos motivated researchers to work on Content-Based Image Retrieval (CBIR).
There are many type of tasks in CBIR: the most studied is the instance-level image search, that consists in retrieving the most similar images starting from an image, used as a query.
This task presents several challenges in terms of accuracy, search time and memory occupancy. Another relevant problem lies in the images themselves, which may present noisy features (\eg, trees, person, cars, ...), different lightning conditions, viewpoints and resolution. 

The image retrieval systems are based on a pipeline usually composed by: extraction of local features from the image, aggregation of the extracted features in a compact representation and retrieval of the most similar images. Initially, the focus was on the feature aggregation step and hence different types of  embeddings were proposed in order to reduce the memory used and obtain a more representative global descriptor. Recently, due to the excellent results obtained in many tasks of computer vision, the deep learning approaches have become dominant also in image retrieval tasks. Particularly, Convolutional Neural Networks (CNNs) are adopted for the feature detection and description phase. They allow to densely extract features from images, that are better than the ones extracted with hand-crafted methods like SIFT \cite{lowe1999object} or SURF \cite{bay2006surf}, because they can catch more details on the images through the high number of convolution layers.  

Following the recent advances, this paper introduces a dense-depth detector applied on CNN codes extracted from InceptionV3 \cite{szegedy2016rethinking} network. This strategy augments the number of features in order to reach higher accuracy with a variant of VLAD \cite{jegouVLAD} descriptors, called locVLAD \cite{maglianiLocVLAD}. It outperforms the previous VLAD implementation on several public benchmarks, thanks also to the use of Z-score normalization \cite{wang2014geometric}.
Furthermore a complete comparison and analysis among the other variants of VLAD is presented.

This paper is organized as follows. Section \ref{ref:rewo} introduces the general techniques used in the state of the art. Section \ref{ref:cnn_codes} reviews the methods used for feature extraction. Next, while Section \ref{ref:DDR} exposes the Dense-Depth Representation,  Section \ref{sec:VLAD} describes VLAD algorithm. Section \ref{ref:res} reports the experimental results on four public datasets: Holidays, Oxford5k, Paris6k and UKB. Finally, concluding remarks are reported.

\section{Related work}\label{ref:rewo}

In the last years, the problem of Content-Based Image Retrieval was addressed in many different ways. The first technique that has been developed, was the Bag of Words (BoW) \cite{sivicBoW}. It is a simple method that reaches good results, but consumes a large amount of memory.
After the development of the BoW approach, researchers tried to overcome its weaknesses and implemented several embedding techniques: Hamming embedding \cite{Holidays}, Fisher Vector \cite{perronnin2010large} and VLAD \cite{jegouVLAD}. 
VLAD \cite{jegouVLAD} is the most used embedding techniques that tries to reduce the dimensionality of features, whilst preserving the recognition accuracy. A VLAD vector is a concatenation of the sum of the difference between the features and the relative closest centers, computed by using K-means clustering. There are many different variants of VLAD presented in the literatur in order to solve the weakness of the VLAD descriptors: CVLAD \cite{CVLAD},  CEVLAD \cite{CEVLAD}, HVLAD \cite{HVLAD}, FVLAD \cite{FVLAD}, gVLAD \cite{wang2014geometric} and locVLAD \cite{maglianiLocVLAD}.
CEVLAD \cite{CEVLAD} applies entropy for the aggreation of the features, FVLAD \cite{FVLAD} modifies the aggregation steps using two codebooks: a descriptor one and a residual one, HVLAD \cite{HVLAD} introduces a hierarchy of codebooks, that allows to create a more robust version of VLAD descriptors.
Then, gVLAD \cite{wang2014geometric} creates different VLAD using the orientation of the features, that are concatenated. This process increases the performance, but requires extra time. 

Recently, with the development of new powerful GPUs, the deep learning approach has shown its superior performance in many tasks of image retrieval. Arandjelovic \etal, in \cite{arandjelovic2016netvlad}, applied a VLAD layer at the end of a CNN architecture, showing that the CNN-based pipeline reaches excellent results in the retrieval task.
Another improvement of deep learning techniques is in the feature extraction phase. This process is known as ``transfer learning'' and consists in tuning the parameters trained in one feature space in order to work in another feature space \cite{pan2010survey}. Some methods that use transfer learning are: Spatial pooling \cite{razavian2016visual}, MOP-CNN \cite{gong2014multi}, Neural codes \cite{babenko2014neural}, Ng \textit{et al.} \cite{yue2015exploiting}, CCS \cite{yan2016cnn}, OC \cite{reddy2015object}, R-MAC \cite{tolias2015particular}, Gordo \textit{et al.} \cite{gordo2016deep} and Magliani \textit{et al.} \cite{magliani2018accurate}. Also, fine-tuning global descriptors \cite{gordo2017end} on a similar image dataset, allows to highly improve accuracy results, but with an extra time effort due to the training phase on the new dataset.

\section{System Architecture} \label{ref:sys_architecture}

In the following subsections, the proposed approach for features extraction and encoding using CNN features and locVLAD embedding is described.

\subsection{CNN codes}\label{ref:cnn_codes}

CNN codes are feature vectors extracted from pre-trained networks using the knowledge gained while solving one problem and applying it to a different, yet related, one through the process of transfer learning. There are several different pre-trained CNN architectures (like VGG16 \cite{simonyan2014very}, GoogLeNet \cite{szegedy2015going}, ResNet \cite{he2016deep}) that allow to easily extract features from their layers. The choice of the layers depends on the type of problem and the selected network. Obviously, the deeper the network is, the better the results obtained by the extracted features are. 



In this paper, the selected network is the recent Inception V3 \cite{szegedy2016rethinking}, because it allows to obtain a more proper representation than VGG16 thanks to the concatenation of different convolutions. From it, the CNN codes in this paper are extracted from the 8th inception pooling layer (called mixed8 in the Keras implementation). Both the network and the layers have been chosen since they achieved the best results in our experiments. An ablation analysis of different networks has been conducted and it is reported in the Section \ref{ref:res}.

\subsection{Dense-Depth representation}\label{ref:DDR}

Since VLAD-based embedding works better with dense representations, we introduced a novel representation scheme.

\begin{figure}[h]
\centering
\includegraphics[width=8cm]{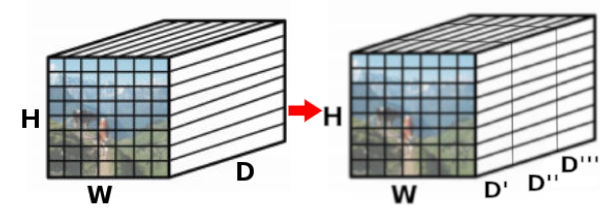}
\caption{Dense-depth detector}
\label{feature_maps}
\end{figure}

The features extracted from mixed8 are grouped into a larger set of features of lower dimensionality in order to augment the VLAD descriptive quality. Given a feature map of dimension $W \times H \times D$ ($W$ = width, $H$ = height and $D$ = depth), we split it along the depth axis in order to obtain a high number of features of lower dimension, as it can be seen in Fig. \ref{feature_maps}.  
Splitting along $D$ allows to mantain the geometrical information of the feature maps, because features that have different position on $H \times W$ are not aggregated.
Following this method, the number of descriptors changes from $H \times W$ to $H \times W \times \frac{D}{split factor}$. 
The $split factor$ indicates the dimension of every single descriptor obtained after the split along the depth axis. This value needs to be a trade-off between the number of features and their discriminative quality.
As an example, a feature map of 8x8x1280, with $split factor = 128$, will be transformed in a set of 8*8*10 = 640 descriptors of 128D. Thereinafter, we will refer to this as \textit{Dense-Depth Representation} (DDR). 

After the extraction, the CNN codes are normalized using the root square normalization described in \cite{arandjelovic2012three}.

We will demonstrate in the experiments that the newly proposed DDR achieves higher performance. 

\subsection{VLAD and locVLAD}\label{sec:VLAD}
Starting from a codebook $C = \{\mathbf{u_1}, \dots, \mathbf{u_K}\}$ of $K$ $d$-dimensional visual features, generated from K-means clustering on the CNN codes, every local descriptor  $X =\{\mathbf{x_1}, \dots, \mathbf{x_m}\}$, extracted from the image, is assigned to the closest cluster center of the codebook:

\begin{equation}
q : \mathbb{R}^d \to C
\end{equation}
\begin{equation}
 q(\mathbf{x}) = \mathbf{\mu_i} \suchthat* i = \argmin_{i = 1, \dots, K} ||\mathbf{x-\mu_i}||
\end{equation}
\noindent where $|| \cdot ||$ is a proper $d$-dimensional distance measure and $d$ is the size of the descriptors ($d$ = 128 in the aforementioned example). 

The VLAD vector $\mathbf{v}$ is obtained by computing the sum of residuals, that is  the difference between the feature descriptor and the corresponding cluster center:
\begin{equation}
 \mathbf{v}^i = \sum\limits_{\forall \mathbf{x} \in X : q(\mathbf{x}) = \mathbf{\mu}_i}{\mathbf{x} - \mathbf{\mu}_i} 
\end{equation}

The last step is the concatenation of the $\mathbf{v}^i$ features, resulting in the unnormalized VLAD vector $\mathbf{v}$.
This vector often includes repeated (duplicated) features that make the recognition more difficult. To solve this problem, the vector $\mathbf{v}$ is normalized using the Z-score normalization \cite{wang2014geometric}, which consists in a residual normalization that is an independent residual $L_2$ norm of every visual word $\mathbf{v}^i$:
\begin{equation}
 \tilde{\mathbf{v}}^i = \sum\limits_{\forall \mathbf{x} \in X : q(\mathbf{x}) = \mathbf{\mu}_i}{\frac{\mathbf{x}-\mathbf{\mu}_i}{||\mathbf{x}-\mathbf{\mu}_i||}} 
\end{equation}
Next, the resulting vector is further normalized as follows: 
\begin{equation}
 \hat{\mathbf{v}}^i = \frac{\tilde{\mathbf{v}}^i - m}{\sigma}
\end{equation}
\noindent where $m$ and $\sigma$ represent the mean and the standard deviation of the vector, respectively. 
Finally, the vector is further $L_2$ normalized.

Recently, with the introduction of locVLAD \cite{maglianiLocVLAD}, the accuracy in retrieval has been increased in an intuitive way. The VLAD vector is calculated through the mean of two different VLAD descriptors: one computed on the whole image and one computed only on the central portion of the image. The idea behind this method is that the most important and useful features in the images are often in the central region. 
LocVLAD is applied only to the query images, because the database images contain different views of the same landmark, as well as zoomed views.

At the end, VLAD descriptors are PCA-whitened \cite{jegou2012negative} to 128D.

\section{Experimental results}\label{ref:res}

The proposed approach has been extensively tested on public datasets in order to evaluate the accuracy against the state of the art. 

\subsection{Datasets and evaluation metrics}

The performance is measured on four public image datasets:
\begin{itemize}
\item Holidays \cite{Holidays} is composed by 1491 images representing the holidays photos of different locations, subdivided in 500 classes. The database images are 991 and the query images are 500, one for every class.

\item Oxford5k \cite{Oxford} is composed by 5062 images of Oxford landmarks. The classes are 11 and the queries are 55 (5 for each class).

\item Paris6k \cite{Paris} is composed by 6412 images of landmarks of Paris, France. The classes are 11 and the queries are 55 (5 for each class).

\item UKB \cite{nister2006scalable} is composed by 10200 images of  diverse categories such as animals, plants,  etc., subdivided in 2550 classes. Every class is composed by 4 images. All the images are used as database images and only one for category is used as a query image.
\end{itemize}


The vocabulary for the VLAD descriptors was created on Paris. Instead, when testing on Paris, the vocabulary was created on Oxford.

For the evaluation of the retrieval performance, mean Average Precision (mAP) was adopted. Instead, for the calculation of the distances between the query VLAD descriptor and the database VLAD descriptors, $L_2$ distance was employed.

In terms of actual implementation, the detector and descriptor used is CNN-based described in section \ref{ref:cnn_codes}, that runs on a NVIDIA GeForce GTX 1070 GPU mounted on a computer with 8-core and 3.40GHz CPU. 
All experiments have been run on 4 threads. To implement the feature extractor system, the Keras library was used. 

\subsection{Results on Holidays}

Table \ref{CNN_res} reports the results obtained extracting the features with CNN  pre-trained on ImageNet \cite{deng2009imagenet}. Different CNN architectures have been tested: VGG19 \cite{simonyan2014very} and InceptionV3 \cite{szegedy2016rethinking}.

All the experiments were executed using a vocabulary of $K=100$ visual words, created on Paris dataset through the application of K-Means \cite{lloyd1982least} clustering technique, initialized following the K-Means++ \cite{ostrovsky2006effectiveness} strategy.
At the beginning, the initial configuration was to extract features from the block4\_pool of VGG19 with an input image of the dimensions equal to 224x224, that is the predefined input image of VGG19. Changing, the dimension of the input image, the results were improved. Also, the choice of the layer in which extract the feature maps is important: from block4\_pool to block5\_pool of VGG19 there was an improvement equal to 2\%.
The breakthorugh was the use of InceptionV3. Thanks to the depth of this CNN architecture, the feature maps extracted allowed to create a more representative VLAD descriptor. The first experiment executed on InceptionV3 reached a mAP equal to 81.55\%, which is almost 4\% more than VGG19.
After we found the best configuration for the parameters: Network and Layer we focused on the other parameters. The application of locVLAD instead of VLAD for the feature aggregation phase allowed to improve the performance (improvement of 3\%). Also, the root square normalization slightly improved the retrieval accuracy. Following the idea that features extracted on the image resized to square not respect the aspect ratio it is not a good idea, we modified the input image of InceptionV3 allowing to have an input of variable size, that was adaptable to the different dimensions of the images.
Finally, the application of PCA-whitening reduced the dimension of the descriptors and removed the noisy values, also improving the performance.

\begin{table*}
\centering
\setlength{\tabcolsep}{3pt}
    \begin{tabular}{|c|c|c|c|c|c|c|c|}
    \hline
     \textbf{Network} & \textbf{Layer} &  \textbf{Input image} & \textbf{DDR} & \textbf{\makecell{Root square \\ norm.}} &  \textbf{Encoding} & \textbf{PCA-whit.} &  \textbf{mAP} \\ \hline
    VGG19 & block4\_pool & 224x244 & \checkmark & \xmark & VLAD & \xmark & 74.33 \\ \hline
    VGG19 & block4\_pool & 550x550 & \checkmark & \xmark & VLAD & \xmark & 75.95  \\ \hline
    VGG19 & block5\_pool & 550x550 & \checkmark & \xmark  & VLAD & \xmark & 77.80  \\ \hline
    InceptionV3 & mixed\_8 & 450x450 & \checkmark & \xmark  & VLAD & \xmark & 81.55  \\ \hline
    InceptionV3 & mixed\_8 & 450x450 & \checkmark & \xmark  & locVLAD & \xmark & 84.55  \\ \hline   
    InceptionV3 & mixed\_8 & 562x562 & \checkmark & \xmark  & locVLAD & \xmark & 85.34  \\ \hline  
    InceptionV3 & mixed\_8  & 562x562 & \checkmark & \checkmark & locVLAD & \xmark & 85.98  \\ \hline  
    InceptionV3 & mixed\_8 & 562x662* & \checkmark & \checkmark   & locVLAD & \xmark & 86.70  \\ \hline 
    InceptionV3 & mixed\_8 & 562x662* & \checkmark & \checkmark & locVLAD & 128D & 87.38 \\ \hline 
    InceptionV3 & mixed\_8 & 562x662* & \xmark & \checkmark  & locVLAD & 128D & 85.63 \\ \hline
    InceptionV3 & mixed\_8 & 562x662* & \checkmark & \checkmark  & locVLAD & 256D & 89.93 \\ \hline
    InceptionV3 & mixed\_8 & 562x662* & \checkmark & \checkmark  & locVLAD & 512D & 90.46 \\ \hline
    \end{tabular}
\caption{Results on Holidays. * indicates cases where the image aspect ratio is mantained.}
\label{CNN_res}
\end{table*}

It is worth to note that all the VLAD and locVLAD vectors are then finally normalized using Z-score normalization.
Furthermore, the application of DDR allowed to higly improve the retrieval performance, as reported in the fourth last row of the Table \ref{CNN_res}.

\subsection{Comparison with the state of the art on Holidays, Oxford5k, Paris6k and UKB}

\begin{table}[htb]
\caption{Comparison of state-of-the-art methods on different public CBIR datasets.}
\centering
\setlength{\tabcolsep}{6pt}
    \begin{tabular}{|c|c|c|c|c|c|}
    \hline
     \textbf{Method} &  \textbf{Dimension} &  \textbf{Oxford5k} & \textbf{Paris6k} & \textbf{Holidays} & \textbf{UKB} \\ \hline
    VLAD \cite{jegouVLAD} & 4096 & 37.80 & 38.60 & 55.60 & 3.18 \\ \hline
    CEVLAD \cite{CEVLAD} & 128 & 53.00 & - & 68.10 & 3.093 \\ \hline
    FVLAD \cite{FVLAD} & 128 & - & - & 62.20 & 3.43 \\ \hline
    HVLAD \cite{HVLAD} & 128 & - & - & 64.00 & 3.40 \\ \hline
    gVLAD \cite{wang2014geometric}  & 128 & \textbf{60.00} & - & 77.90 & -  \\ \hline
    Ng \textit{et al.} \cite{yue2015exploiting}  & 128 & 55.80 & 58.30 & 83.60 & -  \\ \hline 
    \textbf{DDR locVLAD} & 128 & 57.52 & \textbf{64.70} & \textbf{87.38} & \textbf{3.70} \\ \hline
    \hline
    NetVLAD \cite{arandjelovic2016netvlad}  & 512 & 59.00 & 70.20 & 82.90 & -  \\ \hline
    \textbf{DDR locVLAD} & 512 & \textbf{61.46} & \textbf{71.88} & \textbf{90.46} & \textbf{3.76} \\ \hline
    \end{tabular}
\label{CNN_state_of_the_art}
\end{table}

Table \ref{CNN_state_of_the_art} reports the comparison with VLAD approaches on some public datasets.

All the descriptors of the VLAD-based methods are PCA-whitened, except the firt one.
Our method outperforms the others on all the presented datasets, reaching good results on the public benchmarks, in particular on the Holidays dataset.
Unfortunately, on Oxford5k, gVLAD obtained an accuracy slightly better than our due to the concatenation of different VLAD, calculated following the orientation of local features.
On the other hand, augmenting the dimension of the the descriptor from 128D to 512D, our method performs better than the others even on Oxford5k. 
The high accuracy is the result of the focus on the central features of the images due to locVLAD and the dense representation of DDR.


\section{Conclusions}

The paper presents a new Dense-Depth Representation that allows, combined with locVLAD descriptors, to outperform the state of the art related to VLAD descriptors on several public benchmarks. 
The combination of the dense representation (DDR) and the focus on the most important part of the images of locVLAD allowed to obtain a better representation of each image and, therefore, to improve the retrieval accuracy without the need to augment the dimension of the descriptor that still remains 512D.
The future work will be on a different embedding like R-MAC. Also, the application of fine-tuning could help to improve the final accuracy results.

\textbf{Acknowledgment}. This work is partially funded by Regione Emilia Romagna under the “Piano triennale alte competenze per la ricerca, il trasferimento tecnologico e l’imprenditorialità”.


\bibliographystyle{splncs_srt}
\bibliography{egbib}
\end{document}